\pdfoutput=1
\documentclass[a4paper,11pt,notitlepage]{article}

\usepackage[T1]{fontenc}

\usepackage[margin=20mm]{geometry}
\usepackage{setspace}

\usepackage{amsmath, amssymb, bm}
\usepackage{latexsym}

\usepackage{graphicx}
\usepackage{float}
\usepackage{placeins}

\usepackage{booktabs}
\usepackage{multirow}
\usepackage{tabularx}
\usepackage{makecell}
\usepackage{threeparttable}
\usepackage{siunitx}
\usepackage{adjustbox}

\usepackage[table, dvipsnames]{xcolor}

\usepackage{caption}

\usepackage{url}
\usepackage{xspace}

\usepackage[hidelinks]{hyperref}
\usepackage[capitalise]{cleveref}

\usepackage{soul}

\usepackage{enumitem}

\usepackage{listings}
\usepackage{algorithm}
\usepackage[noend]{algpseudocode}
\usepackage{tikz}
\usetikzlibrary{positioning,backgrounds,fit}

\tikzset{
  bubble/.style = {
    draw=black!25,
    fill=white,
    rounded corners,
    align=left,
    font=\footnotesize,
    line width=0.3pt,
    inner sep=6pt,
    text width=\dimexpr\textwidth-12pt\relax
  },
  bubbleAlt/.style = {bubble, fill=black!3},
}

\usepackage[most]{tcolorbox}
\tcbset{
  colback=gray!10,
  colframe=gray!60!black,
  boxrule=0.4pt,
  arc=2pt,
  left=6pt,right=6pt,top=6pt,bottom=6pt
}
\newtcolorbox{codebookbox}[1][]{breakable,#1}

\newcommand{\keywords}[1]{\textbf{\textit{Keywords:}} #1}

\usepackage{cite}
\begin{document}
%\linenumbers
 \begin{center}
   \LARGE{\textbf{LinGO: A Linguistic Graph Optimization Framework with LLMs for Interpreting Intents of Online Uncivil Discourse}}
 
\vspace{1cm}

\large Yuan Zhang$^{1,\ast}$, Thales Bertaglia$^{2}$

\vspace{0.2cm} \normalsize 
$^{1}$University of Zurich, Zurich, Switzerland\\
$^{2}$Utrecht University, Utrecht, Netherlands\\
$\ast$ y.zhang@ikmz.uzh.ch

\end{center}

\begin{abstract}
Detecting uncivil language is crucial for maintaining safe, inclusive, and democratic online spaces. Yet existing classifiers often misinterpret posts containing uncivil cues but expressing civil intents, leading to inflated estimates of harmful incivility online. We introduce LinGO, a linguistic graph optimization framework for large language models (LLMs) that leverages linguistic structures and optimization techniques to classify multi-class intents of incivility that use various direct and indirect expressions. LinGO decomposes language into multi-step linguistic components, identifies targeted steps that cause the most errors, and iteratively optimizes prompt and/or example components for targeted steps. We evaluate it using a dataset collected during the 2022 Brazilian presidential election, encompassing four forms of political incivility: Impoliteness (IMP), Hate Speech and Stereotyping (HSST), Physical Harm and Violent Political Rhetoric (PHAVPR), and Threats to Democratic Institutions and Values (THREAT). Each instance is annotated with six types of civil/uncivil intent. We benchmark LinGO using three cost-efficient LLMs: GPT-5-mini, Gemini 2.5 Flash–Lite, and Claude 3 Haiku, and four optimization techniques: TextGrad, AdalFlow, DSPy, and Retrieval-Augmented Generation (RAG). The results show that, across all models, LinGO consistently improves accuracy and weighted F1 compared with zero-shot, chain-of-thought, direct optimization, and fine-tuning baselines. RAG is the strongest optimization technique and, when paired with Gemini model, achieves the best overall performance. These findings demonstrate that incorporating multi-step linguistic components into LLM instructions and optimize targeted components can help the models explain complex semantic meanings, which can be extended to other complex semantic explanation tasks in the future.

\end{abstract} 
\section{Introduction}

Online incivility is a multidimensional concept incorporating behaviors that are impolite, aggressive, harmful, and even detrimental to democratic values, thereby undermining online safety and democratic processes \cite{muddiman2017incivility, stryker2016political, bentivegna2022incivility}. Both computer scientists and social scientists have sought to identify uncivil discourse online using Artificial Intelligence (AI) techniques \cite{stoll2020detecting, sadeque2019incivility, gaydhani2018detecting, nugroho2019improving, chakrabarty2019machine, davidson2020incivility, ali2022hate, mozafari2019bert, abusaqer2025efficient, gao2024crisis}. Although these machine learning and deep learning methods enable large-scale classification, post hoc analyses show frequent errors and interpretive difficulty, even for human coders, when dealing with many indirect expressions. One common case is implicit attacks that rely on sarcasm, insinuation, presupposition, or metaphor, or that require inferring incivility from additional context \cite{madhyastha2023contextual}. This type of incivility is not perceived equally by everyone and is often considered milder than explicit attacks \cite{bormann2022perceptions}. For instance, "Brilliant leadership, if the goal was to embarrass the nation" expresses dissatisfaction and criticism of the leadership but does not really contain harmful intent \footnote{The example is generated by AI.}. A second case involves referencing others’ uncivil speech while adding the author’s own opinions, such as "They told us to go back where we came from. This is unacceptable." \footnote{The example is generated by AI.} The author’s stance toward the quoted content is condemning and it implies an intent of counterspeech. When the author’s stance is endorsing or neutral/reporting, the conveyed intent may shift, for example, toward amplification or just attention-seeking. Treating all discourse that contains uncivil elements as having the same harmful intent, as earlier AI models largely did, risks overestimating the prevalence of incivility and harm online \cite{davidson2017automated, rottger2020hatecheck, vanaken2018challenges}. 

LLMs, drawing on knowledge from large-scale training data, can capture more nuanced semantic meaning than earlier AI models, but they remain struggle to recognize complex linguistic structures in zero-shot settings \cite{andersson2025chatgpt_impoliteness}. With recent advances of LLM optimization techniques, we can leverage LLMs more effectively by automatically programming the prompts and/or few-shot examples, similar to how traditional neural networks are trained \cite{khattab2023dspy, yuksekgonul2024textgrad, yin2025llm_autodiff}. Additionally, many linguistic structures follow systematic patterns and convey consistent intent. Incorporating multiple linguistic structural components into LLM instructions could help LLMs identify and optimize specific comprehension difficulties and improve their overall interpretive ability. Therefore, this work introduces a linguistic graph optimization framework that integrates linguistic structures with existing LLM optimization techniques and improves performance in recognizing various intents of uncivil language uses.

\begin{figure}[t]
\centering

% Scale down only if needed (keeps aspect ratio)
\begin{adjustbox}{max width=\textwidth,center}
\begin{tikzpicture}[node distance=0.8cm]

% --- set bubble width for EACH column ---
\def\bubbleW{0.40\textwidth}  % tweak: 0.38–0.42 usually best in LNCS

% -------- Left column (1–3) --------
\node (Limg1) {\includegraphics[width=0.85cm]{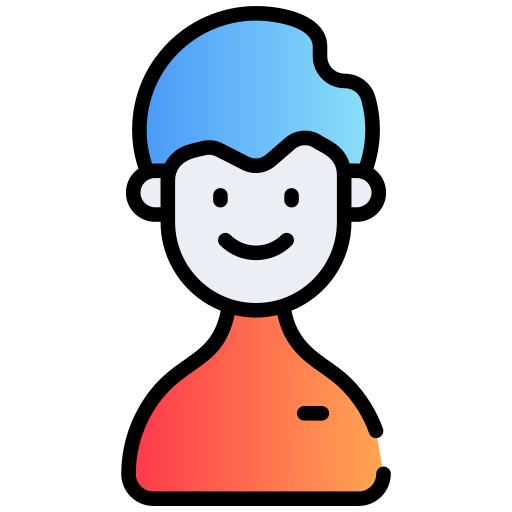}};
\node[bubble, text width=\bubbleW, right=0.35cm of Limg1] (L1)
  {{\bfseries [1] Explicit}\par ``Kill all the [group].''};

\node (Limg2) [below=of Limg1] {\includegraphics[width=0.85cm]{user.png}};
\node[bubbleAlt, text width=\bubbleW, right=0.35cm of Limg2] (L2)
  {{\bfseries [2] Implicit}\par ``We all know what [group] are like.''};

\node (Limg3) [below=of Limg2] {\includegraphics[width=0.85cm]{user.png}};
\node[bubble, text width=\bubbleW, right=0.35cm of Limg3] (L3)
  {{\bfseries [3] Reporting}\par ``He said, ` [group] are ruining this country.' ''};

% -------- Right column (4–6) --------
\node (Ranchor) [right=0.9cm of L1] {}; % reduce gap if still too wide

\node (Rimg1) [right=0cm of Ranchor] {\includegraphics[width=0.85cm]{user.png}};
\node[bubbleAlt, text width=\bubbleW, right=0.35cm of Rimg1] (R1)
  {{\bfseries [4] Intensifying}\par ``He's right—[group] are a plague.''};

\node (Rimg2) [below=of Rimg1] {\includegraphics[width=0.85cm]{user.png}};
\node[bubble, text width=\bubbleW, right=0.35cm of Rimg2] (R2)
  {{\bfseries [5] Criticizing}\par ``It's wrong to say ` [group] are animals.' ''};

\node (Rimg3) [below=of Rimg2] {\includegraphics[width=0.85cm]{user.png}};
\node[bubbleAlt, text width=\bubbleW, right=0.35cm of Rimg3] (R3)
  {{\bfseries [6] Escalating}\par ``You call [group] parasites, but your kind is even worse.''};

% -------- Background frame --------
\begin{pgfonlayer}{background}
  \node[
    fill=black!2,
    draw=black!20,
    rounded corners,
    inner sep=6pt,
    fit=(Limg1) (L1) (Limg3) (L3) (Rimg1) (R1) (Rimg3) (R3)
  ] {};
\end{pgfonlayer}

\end{tikzpicture}
\end{adjustbox}

\caption{Illustration of six intent categories of hate speech. [1] is a direct expression of hate speech. [2]--[6] are indirect expressions of hate speech.}
\label{fig:intent-dialogue-pro}
\end{figure}

In summary, this work addresses a multi-label classification task that distinguishes direct and indirect expressions of incivility, which are closely tied to speakers’ intents. This distinction is important because not all intents are equally harmful; given this classification, future studies could examine the harmfulness of each category in greater detail. Fig.~\ref{fig:intent-dialogue-pro} presents six expression ways in which hate speech indicators are present but intents vary. Cases \([1]\) and \([2]\) are hateful opinion from the author toward a protected group: \([1]\) is explicit and overtly uncivil, whereas \([2]\) is implicit and subtler, often lacking obvious surface cues and harmful intent. By contrast, Cases \([3]\)–\([6]\) reference a third party’s hateful statement; we further consider the author’s stance: In \([5]\), the author criticizes or rejects the hateful content, yielding counter-speech of incivility. With no stance (pure reference), agreement, and response with additional hate speech from the author, the intent shifts to reporting \([3]\), intensifying \([4]\), and escalating \([6]\) others' hate speech. Under these circumstances, only case \([1]\) constitutes direct harmful intent, whereas the other cases involve indirect expressions that are not intended to cause harm but may still be harmful depending on the specific context.

Those complex expressions are highly structured and require multi-step comprehension for LLMs, similar to how human beings cognitively process them. Therefore, we hypothesize that a multi-step procedure guided by linguistic structure could also help LLMs capture the intended meaning more effectively \cite{rottger2020hatecheck}. For example, reference of incivility can be modeled as a nested construction:
[\textit{author’s stance} [\textit{reference to others’ statements/\allowbreak behaviors}]].
First, we detect incivility within the [\textit{reference to others’ statements/\allowbreak behaviors}].
If incivility is present, treat the instance as \emph{indirect} and then assess the author’s stance
in [\textit{author’s stance}]. If no incivility is found in the reference, treat the instance as
\emph{direct} and evaluate incivility in [\textit{author’s stance}] alone. When no reference appears,
set [\textit{reference to others’ statements/\allowbreak behaviors}] = \textit{None} and detect
incivility solely within [\textit{author’s stance}]. Although similar idea has before been put forward as functionalities checklist \cite{rottger2020hatecheck} or as an analytical scheme \cite{vanaken2018challenges}, it has not yet been systematically programmed with AI models, hence large-scale structural linguistic understanding remains underexplored. To fill this gap, we incorporate linguistic components into LLM-based inference in this work. Specifically, we use a linguistic graph in which the output of each step determines the next step, until the path reaches the final decision label (see more details in Method).

This multi-step linguistic graph is further combined with current automated optimization techniques for improving stability. As is well known, LLM outputs are highly sensitive to the prompt and to any included examples (in few-shot settings); even small changes can significantly affect performance. This disadvantage is amplified in multi-step tasks due to error propagation: a mistake made in an early step can carry forward, causing downstream steps to fail or make further errors, often magnifying the original mistake \cite{caselli2015pilingup}. The simplest strategy overcoming the instability is manual trial-and-error, where practitioners iteratively refine prompts after observing misinterpretations \cite{sahoo2024survey}. However, this strategy requires substantial human effort and relies excessively on subjective decisions. Scholars have therefore proposed systematic methods that treat prompt/example optimization as a programming task \cite{li2025automatic}. Existing techniques include textural gradient descent (e.g., TextGrad, AdalFlow), similarity-based retrieval (e.g., retrieval-augmented generation, RAG), metric-driven bootstrapping (e.g., DSPy), etc. Combining the linguistic graph with optimization techiniques, our approach enables to identify the most problematic steps during sentence comprehension, and optimization techniques can develop the best strategies for those steps. In this way, error propagation can be alleviated.

The whole procedure is summerized as Linguistic Graph Optimization. First, we construct a graph of linguistic nodes grounded in linguistic structure. Second, the LLM traverses this graph step by step, where each next step depends on the output of the previous one. Finally, the model optimizes prompts and/or examples for the steps with the most frequent errors. 

The results of our experiments show that, for all models, stronger baselines that incorporate some automated optimization techniques outperform basic baselines such as zero-shot prompting and Chain-of-Thought (CoT) prompting, even without applying linguistic graph. Among these techniques, RAG achieves the highest weighted F1 scores (OpenAI: 0.640; Gemini: 0.664; Claude: 0.551), suggesting that optimizing demonstration examples through high-quality selection may be more effective than text-based optimization. Moreover, except for the Claude model, incorporating the linguistic-graph multi-step procedure with both OpenAI and Gemini models yields further improvements across all optimization techniques. Overall, the best performance is achieved by RAG with the Gemini model, reaching 0.690 accuracy and a 0.699 weighted F1 score. We also compare our approach’s best performance with that of open-source fine-tuning, and it still achieves the best results. Our findings indicate that decomposing complex semantic understanding into multiple steps and identifying key steps for optimization helps improve the interpretability of LLMs.

\section{Related Work}

\textbf{Automatic Identification of Uncivil Discourses.}
Incivility refers to behaviors that violate norms of politeness or democratic values \cite{muddiman2017incivility, stryker2016political}. Natural Language Processing (NLP), leveraging machine learning and deep learning techniques, enables scalable automatic detection of such online discourse \cite{seble2023hate}.

Early works mostly relied on classical statistical machine learning. These approaches typically compute term frequency–inverse document frequency (TF–IDF) features or use bag-of-words (BoW) representations, which are then fitted with models such as Naive Bayes, Logistic Regression, Random Forests, and Support Vector Machines (SVMs) \cite{stoll2020detecting, sadeque2019incivility, gaydhani2018detecting, nugroho2019improving, chakrabarty2019machine}. While these methods effectively detect straightforward cases with clear lexical indicators, they perform poorly when incivility is subtle or context-dependent \cite{stoll2020detecting}

With the advent of neural network and pre-trained embeddings, performance on complex forms of incivility has substantially improved \cite{mozafari2019bert}. Unlike frequency-based or purely probabilistic word representations, embeddings capture relational and semantic properties in dense vector spaces \cite{mikolov2013efficient}. Concurrent advances in neural architectures, such as Recurrent Neural Networks (RNNs), Long Short-Term Memory (LSTM) networks, and Transformers, have enabled richer modeling of context \cite{elman1990finding, hochreiter1997long, vaswani2017attention}. Numerous studies using transformer-based models, particularly BERT and its derivatives, report strong performance for incivility detection, including Davidson et al. \cite{davidson2020incivility}, Ali et al. \cite{ali2022hate}, Mozafari et al. \cite{mozafari2019bert}, Abusaqer et al. \cite{abusaqer2025efficient}, and Gao \cite{gao2024crisis}, among others. 

\textbf{Indirect Incivility and Linguistic Structure.}
Despite substantial progress of artificial intelligence technologies, automatic incivility detection remains far from solved. One major challenge is indirect incivility: texts that contain uncivil elements but express them implicitly (i.e., through sarcasm or metaphor, by obscuring context), or by referencing others’ uncivil statements with additional commentary or opinions\cite{davidson2017automated, rottger2020hatecheck, gligoric2024nlp}. For example, Zhang et al. \cite{zhang2025quantifying} used multilingual sentence transformers to identify multidimensional forms of political incivility and, upon manual inspection, found many indirect cases identified by classifiers, particularly those involving physical harm and threats to democratic values. Implicit incivility is difficult even for human coders to judge, and it often leads to low-quality annotations \cite{albladi2025hate_speech_llm_review}. People sometimes also cite others’ uncivil statements or behaviors to report it, criticize it, or escalate the discussion, and their intents differ from directly expressing incivility. To interpret their intents and avoid exaggeration of harmfulness, models must capture not only superficial uncivil features but also their linguistic structures \cite{rottger2020hatecheck}.

Using linguistically informed tests to probe indirect cases of incivility is not a novel idea. For instance, Röttger et al. present a diagnostic suite of 29 linguistically motivated cases to stress-test classifiers \cite{rottger2020hatecheck}. Van Aken et al. conduct an in-depth linguistic error analysis of misclassified toxic comments \cite{vanaken2018challenges}. Although these studies illuminate the important role of linguistic structure in identifying indirect expressions, to our knowledge this has not yet been incorporated into the programming of AI models. Incorporating linguistic rules into the LLM instructions is likely to improve LLMs’ understanding of indirect and complex semantics.

\textbf{LLMs with Automatic Optimization.}
Trained on large corpora and optimized under an autoregressive objective, generative LLMs can produce high-quality outputs across a wide range of tasks. In a common setup, the natural language prompt and few-shot examples are tokenized, and the model generates an output by sequentially predicting the next tokens \cite{liu2023pretrain}. However, response quality depends heavily on the prompts and examples provided. A popular direction is to use automatic optimization techniques to help humans find effective prompts and/or examples with minimal effort \cite{shin2020autoprompt}. Current widely used techniques include: (i) gradient-based methods that treat prompts or examples as trainable parameters and optimize them via gradient descent \cite{shin2020autoprompt,pryzant2023automatic}. For instance, TextGrad introduces a gradient-based optimization scheme for LLM systems in which prompt text (and potentially intermediate reasoning and outputs) is updated using textual gradients generated by a stronger model, analogous to backpropagation in neural network training \cite{yuksekgonul2024textgrad}. Similarly, AdalFlow also uses LLM-generated feedback as a gradient-like signal, but within a broader engineering framework where multiple components, including prompt text, selected examples, and other modules, can be optimized \cite{yin2025llm_autodiff}. (ii) retrieval of representative demonstrations from additional sources. Including representative few-shot examples typically improves LLM performance \cite{brown2020language}, akin to providing supervised signals that the model can generalize to unseen inputs. To retrieve relevant examples, one can borrow techniques from RAG, which selects the most relevant information by similarity matching \cite{lewis2020rag}. In broader applications, RAG can retrieve not only static examples but also up-to-date knowledge from online and offline corpora. (iii) metric-driven prompt construction via bootstrapping. Rather than hand-crafting prompts or manually adding demonstrations, frameworks such as DSPy define task signatures and construct prompts with demonstrations during compilation by bootstrapping examples that optimize user-defined metrics \cite{khattab2023dspy}. DSPy can also be extended to multi-step tasks by composing multiple declarative modules and optimizing them toward a shared objective or module-specific objectives \cite{khattab2023dspy}. Although tools like TextGrad, AdalFlow, and RAG are not intrinsically designed for multi-step tasks, they can also be applied to individual steps or subsets of steps. These programming techniques can be combined with our linguistic graph to improve multi-step reasoning and reduce error propagation.

\section{Method}
The goal of our approach is to classify six direct and indirect expression patterns and intents of incivility: (1) explicit attack (direct); (2) implicit attack (indirect); (3) reporting incivility (indirect); (4) intensifying incivility(indirect); (5) countering incivility (indirect); and (6) escalating incivility (indirect). (0) is a backup label that collects cases that do not belong to any of the categories above. We first define a linguistic graph that incorporates Q-A nodes related to the nested components [\textit{author’s stance} [\textit{reference to others’ statements/\allowbreak behaviors}]], as well as judgments about whether each component contains uncivil elements and whether those elements are expressed explicitly or implicitly. The model’s response at each node determines which next node to visit, until a final label is produced. Furthermore, the framework identifies which specific nodes lead to the most frequent errors by LLMs, enabling targeted step optimization. A detailed description is provided below, and a formal mathematical definition of the procedures are provided in Appendix B.

\subsection{Linguistic Graph Construction}

The linguistic graph consists of five steps and is described in the following and in Algorithm~\ref{alg:lingo}. An illustrative visualization of the linguistic graph is provided in Fig.~\ref{fig:linguistic graph} in the Appendix B. 

\begin{enumerate}[label=\emph{Step \arabic*:}, leftmargin=*, itemsep=0.25em]
  \item Determine whether the post refers to another person’s statement or behavior. If \emph{YES}, go to Step 2; if \emph{NO}, go to Step 4.
  \item Analyze whether the referenced statement or behavior contains explicit or implicit incivility. If \emph{YES}, go to Step 3; if \emph{NO}, go to Step 4.
  \item Examine whether the author reports, intensifies, counters, or escalates the referenced incivility. Assign labels \emph{Report (3)}, \emph{Intensify (4)}, \emph{Counter (5)}, and \emph{Escalate (6)}.
  \item Determine whether the author’s own statement or behavior contains explicit or implicit incivility. If \emph{YES}, go to Step 5; if \emph{NO}, assign label \emph{Other (0)}.
  \item Classify the author’s incivility as explicit or implicit. \emph{Explicit} incivility refers to expressions with salient uncivil features, while \emph{implicit} incivility covers cases without salient features that nonetheless convey uncivil meaning (e.g., critical, sarcastic, or metaphorical). Assign labels \emph{Explicit (1)} and \emph{Implicit (2)}.
\end{enumerate}

\makeatletter
\newcommand{\algmarginshrink}{%
  \setlength{\ALG@thistlm}{\parindent}%
}
\makeatother

\begin{algorithm}[h]
% \captionsetup{font=small}
\caption{Linguistic Graph}
\label{alg:lingo}
\algmarginshrink
\footnotesize
\begin{algorithmic}[1]
\algrenewcommand\algorithmicrequire{\textbf{Input:}}
\algrenewcommand\algorithmicensure{\textbf{Output:}}
\Require \begin{minipage}[t]{0.95\linewidth}Post $x$\end{minipage}
\Ensure \begin{minipage}[t]{0.95\linewidth}
$y \in \{$Other (0), Explicit (1), Implicit (2), Report (3), Intensify (4), Counter (5), Escalate (6)$\}$
\end{minipage}

\Function{ClassifyIntent}{$x$}
  \State $r \gets \Call{DetectReference}{x}$ \Comment{Step 1}
  \If{$r$}
    \State $c \gets \Call{ExtractReferencedContent}{x}$
    \State $h \gets \Call{ContainsIncivility}{c}$ \Comment{Step 2}
    \If{$h$}
      \State $s \gets \Call{DetermineStance}{x, c}$ \Comment{Step 3}
      \If{$s=\text{report}$} \State \Return \textbf{Report (3)} \EndIf
      \If{$s=\text{intensify}$} \State \Return \textbf{Intensify (4)} \EndIf
      \If{$s=\text{counter}$} \State \Return \textbf{Counter (5)} \EndIf
      \State \Return \textbf{Escalate (6)}
    \EndIf
  \EndIf
  \State $o \gets \Call{ExtractAuthorText}{x}$ 
  \State $h_o \gets \Call{ContainsIncivility}{o}$ \Comment{Step 4}
  \If{\textbf{not} $h_o$}
    \State \Return \textbf{Other (0)}
  \Else
    \State $t \gets \Call{TypeExplicitVsImplicit}{o}$ \Comment{Step 5}
    \If{$t=\text{explicit}$} \State \Return \textbf{Explicit (1)}
    \Else \State \Return \textbf{Implicit (2)} \EndIf
  \EndIf
\EndFunction
\end{algorithmic}
\end{algorithm}

\subsection{Human Annotation of Training Data}
Human coders are required to annotate the texts following the steps described in the linguistic graph above. The annotation includes an answer for each sub-step as well as a final classification label. All answers and labels are concatenated into a reasoning chain in the form: \lstinline|STEP 1: Answer -> STEP 2: Answer -> ... -> LABEL: 0-6|. Finally, an intercoder reliability test is conducted to assess the consistency of the annotated sub-step answers and labels. Because the task involves multiple classes and the class distribution is highly imbalanced, we recommend using Gwet’s AC2 as the reliability metric. Compared with traditional measures such as Krippendorff’s alpha, Gwet’s AC2 is less affected by class imbalance becuase of using a more stable model for estimating chance agreement \cite{gwet2008highagreement,gwet2001handbookirr}.

With the annotated data, we can then split the dataset into developing (which is further split into training and validation sets) and test sets following traditional supervised learning practice.

\subsection{Targeted Step Optimization}

The automatic prompt-optimization procedure comprises three steps. 
First, we provide an initial prompt that includes: (i) definitions of specific forms of incivility (can be single category or multi-category); (ii) the overall task description; (iii) questions and instructions at each step defined in the linguistic graph; and (iv) the required output format. We require the model to return both a label and a reasoning chain similar to human being's annotation, e.g.,
\begin{lstlisting}[breaklines=true, basicstyle=\ttfamily\small]
REASONING: [STEP 1: YES/NO -> STEP 2/4: YES/NO ->
STEP 3/5: Other/Explicit/Implicit/Report/Intensify/Counter/Escalate]
-> LABEL: 0--6
\end{lstlisting}

Second, we use generative LLMs to generate responses for data in validation set. The model responses are compared against human-labeled sub-steps, and we examine the distribution of mismatches across all steps. Steps whose proportion of mismatches exceeds certain thresholds are selected for optimization. Optimization is conducted utilizing the existing advanced LLM programming tools: TextGrad, AdalFlow, DSPy, and RAG, and examples from training set. The specific techniques and optimized elements are shown in Tab.~\ref{tab:opt_tools}.

\begin{table}[!htbp]
\centering
\caption{Tools used for step-wise optimization and the corresponding optimized elements.}
\label{tab:opt_tools}
\small
\setlength{\tabcolsep}{4pt}
\begin{tabular}{p{0.14\columnwidth} p{0.24\columnwidth} p{0.56\columnwidth}}
\hline
\textbf{Tool} & \textbf{Technique} & \textbf{Optimized elements} \\
\hline
TextGrad & Textual gradient descent & Targeted prompt text \\
AdalFlow & Textual gradient descent & Targeted prompt text; targeted demonstrations \\
DSPy & Metric-driven bootstrapping & Targeted demonstrations; targeted traces \\
RAG & Similarity-based retrieval & Targeted demonstrations \\
\hline
\end{tabular}
\end{table}

\begin{figure*}[h]
  \centering
  \includegraphics[width=\linewidth]{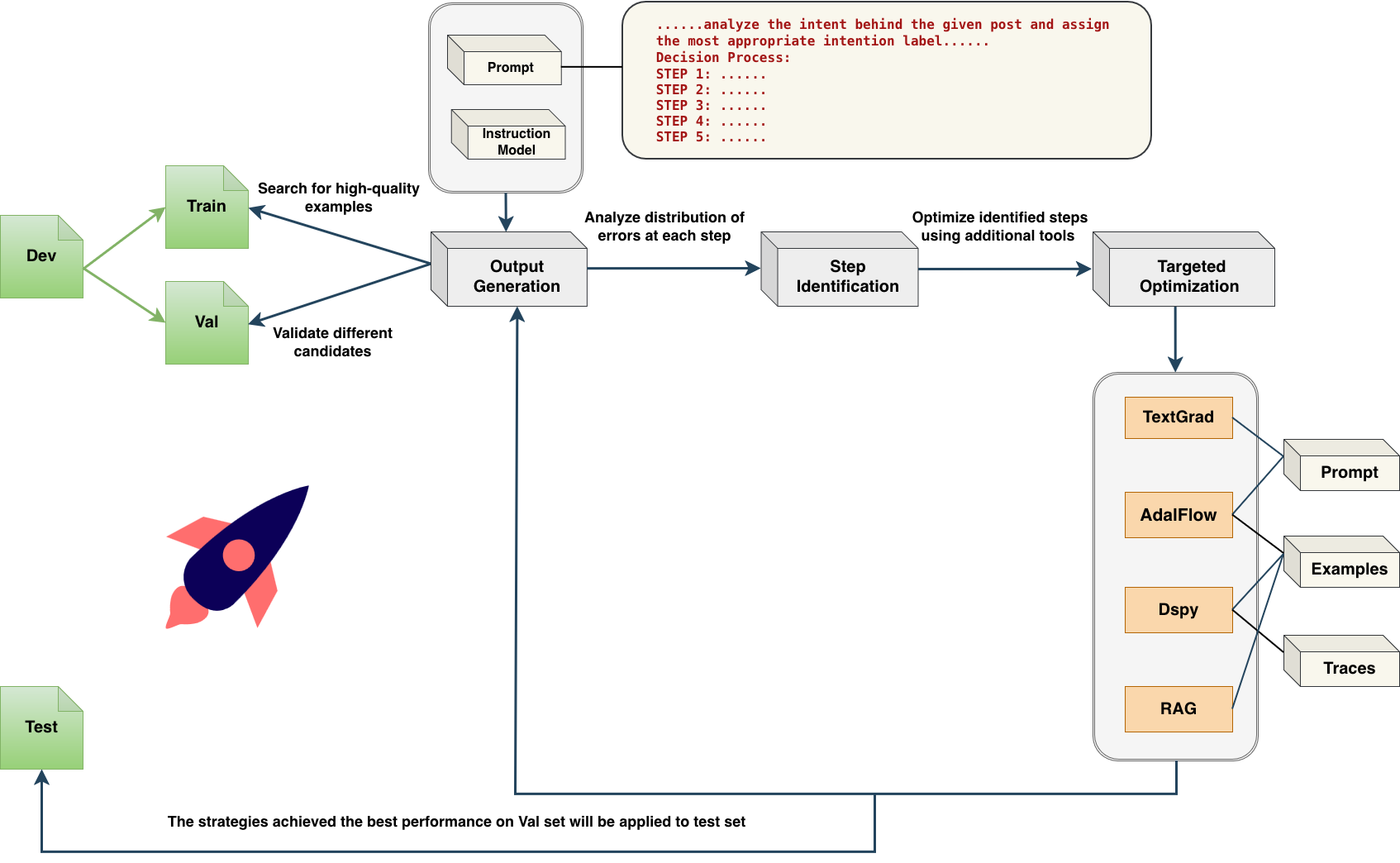} 
  \caption{Demonstration of pipeline of Linguistic Graph Optimization (LinGO).}
  \label{fig:LinGO_pipeline}
\end{figure*}

The optimization procedure can be run iteratively over multiple rounds, and the prompt and/or examples that achieve the best validation performance will then be evaluated on the test set. The algorithmic description is provided in Algorithm~\ref{alg:optimization}. The whole pipeline is visualized in Fig.~\ref{fig:LinGO_pipeline}.

\begin{algorithm}[h]
% \captionsetup{font=small}
\caption{Targeted Optimization}
\label{alg:optimization}
\footnotesize
\begin{algorithmic}[1]
\algrenewcommand\algorithmicrequire{\textbf{Input:}}
\algrenewcommand\algorithmicensure{\textbf{Output:}}
\Require \begin{minipage}[t]{0.96\linewidth}
$\mathcal{T}=(\mathcal{N},\mathcal{E})$ (linguistic graph with steps/nodes $\mathcal{N}$), 
$\mathcal{D}=\mathcal{D}_{\text{train}}\cup\mathcal{D}_{\text{val}}\cup\mathcal{D}_{\text{test}}$ with sub-step labels $a_n^\ast(x)$ and final label $y^\ast(x)$,\\
initial prompt $\mathcal{P}^{(0)}$ (definitions, task description, step instructions, output format), threshold $\tau$, few-shot size $k$, max rounds $R$,\\
LLM $g(\cdot)$, validation metric $m_{\text{val}}(\cdot)$, optimizer set $\Omega=\{\textsc{TextGrad},\textsc{AdalFlow},\textsc{DSPy},\textsc{RAG}\}$
\end{minipage}
\Ensure \begin{minipage}[t]{0.96\linewidth}
Optimized prompt and examples $(\mathcal{P}^\star, \mathcal{F}^\star)$ for final test evaluation
\end{minipage}

\State $\mathcal{P} \gets \mathcal{P}^{(0)}$;\quad $\mathcal{F}\gets \varnothing$;\quad $s^\star \gets -\infty$;\quad $(\mathcal{P}^\star,\mathcal{F}^\star)\gets(\mathcal{P},\mathcal{F})$

\For{$t \gets 0$ \textbf{to} $R-1$}
  \State \textbf{(Run)} For each $(x_i,\cdot)\in \mathcal{D}_{\text{val}}$, query $g$ with $(\mathcal{P},\mathcal{F},x_i)$ and parse
  \Statex \hspace{1.6em}predicted reasoning path $\pi(x_i)$, node answers $\hat a_n(x_i)$, and predicted label $\hat y(x_i)$.

  \State \textbf{(Diagnose)} For each node $n\in\mathcal{N}$, define the visited set 
  \Statex \hspace{1.6em}$\mathcal{I}_n \gets \{\, i:\; n\text{ is visited in }\pi(x_i)\,\}$,
  and compute step-wise mismatch rate
  \Statex \hspace{1.6em}$\hat p_n \gets \dfrac{1}{|\mathcal{I}_n|}\sum_{i\in \mathcal{I}_n}\mathbf{1}\!\left\{\,\hat a_n(x_i)\neq a_n^\ast(x_i)\,\right\}$.

  \State \textbf{(Select)} $\mathcal{S} \gets \{\, n\in\mathcal{N}:\; \hat p_n > \tau \,\}$ \Comment{steps to optimize}
  \If{$\mathcal{S} = \varnothing$} \textbf{break} \EndIf

  \ForAll{$n \in \mathcal{S}$}
    \State \textbf{(Collect errors)} $\mathcal{E}_n \gets \{\, (x_i, a_n^\ast(x_i), \hat a_n(x_i)):\; i\in\mathcal{I}_n \wedge \hat a_n(x_i)\neq a_n^\ast(x_i)\,\}$
    \State \textbf{(Sample)} $\mathcal{F}_n \gets \textsc{Sample}(\mathcal{E}_n, D_\text{train})$ 
    \State \textbf{(Choose tool)} $\omega_n \gets \textsc{SelectOptimizer}(\Omega, n, \mathcal{F}_n)$
    \State \textbf{(Optimize)} $(\Delta \mathcal{I}_n, \Delta \mathcal{F}_n) \gets \omega_n(\mathcal{P}, \mathcal{F}_n, \mathcal{T})$
    \State \textbf{(Update)} $\mathcal{P} \gets \textsc{EmbedNodeUpdate}(\mathcal{P}, n, \Delta \mathcal{I}_n)$
    \Statex \hspace{1.6em}$\mathcal{F} \gets \textsc{ExamplesUpdate}(\mathcal{F}, \Delta \mathcal{F}_n)$
  \EndFor

  \State \textbf{(Validate)} $s \gets m_{\text{val}}(\mathcal{P},\mathcal{F},\mathcal{D}_{\text{val}})$
  \If{$s > s^\star$}
    \State $s^\star \gets s$;\quad $(\mathcal{P}^\star,\mathcal{F}^\star)\gets(\mathcal{P},\mathcal{F})$
  \EndIf
\EndFor

\State \textbf{(Test once)} Evaluate $(\mathcal{P}^\star,\mathcal{F}^\star)$ on $\mathcal{D}_{\text{test}}$ to report final results.
\State \Return $(\mathcal{P}^\star,\mathcal{F}^\star)$
\end{algorithmic}
\end{algorithm}

\subsection{Baseline Approaches as Comparison}

We compare the classification performance of our approach against four baselines: zero-shot prompting, CoT prompting, direct optimization, and LoRA fine-tuning. In the zero-shot setting, the prompt contains only the task description, and the model outputs a label directly without a reasoning chain. In the CoT setting, the prompt includes the same task description but explicitly instructs the model to reason step by step before producing an answer. 

As more advanced baselines, direct optimization updates the prompt and/or examples using different optimization techniques but does not involve the linguistic graph reasoning steps. We also include LoRA fine-tuning as an additional baseline, since it directly updates model parameters and can improve performance while requiring far fewer trainable parameters than full fine-tuning \cite{hu2022lora}.

\section{Experiments}
\subsection{Experimental Setup}

\paragraph{Data Preparation.}  
We use a Portuguese dataset of Twitter/X posts published by political influencers during the 2022 Brazilian presidential election, provided by \cite{zhang2025quantifying} \footnote{Data license available at \url{https://doi.org/10.7910/DVN/M552GM}.}. The dataset contains only publicly available posts and was collected in compliance with Twitter/X’s Terms of Service. Our reuse of the dataset aligns with the original access conditions and intended research purpose. In the paper of using this dataset, posts have been annotated with binary labels (\textit{civil} vs. \textit{uncivil}) across four forms of incivility in political contexts: \textit{Impoliteness}, \textit{Physical Harm and Violent Rhetoric}, \textit{Hate Speech and Stereotyping}, and \textit{Threats to Democratic Institutions and Values}. From each dimension, we randomly sample 500 posts classified as ``uncivil'' by their sentence-transformer models. Two researchers with expertise in the Brazilian context and proficiency in Portuguese then annotate the intents of these posts according to the linguistic graph, providing both the final labels and sub-step answers as ground-truth references. 

Inter-coder agreement is 67.35\%, and inter-coder reliability, measured by Gwet's AC2, is 0.514. We consider this level acceptable given the multi-class complexity of the task.

\paragraph{Model Selection.}  
We select three instruction-tuned LLMs as \textit{instruction models} to generate the initial responses for comparison across commercial providers. Since our goal is to compare improvements across methods rather than the intrinsic capabilities of the underlying models, we use relatively low-cost models to reduce experimental costs. Specifically, we use: (i) GPT-5-mini (OpenAI), (ii) Gemini 2.5 Flash-Lite (Google), and (iii) Claude 3 Haiku (Anthropic). For optimization frameworks that require a teacher or optimizer model, we use GPT-5-mini in all experiments.
For the open-source models used for fine-tuning, we select Mistral-7B, DeepSeek-V2-Lite-Chat, and Qwen3-4B-Instruct-2507 due to their strong instruction-following capabilities and manageable parameter sizes.

\paragraph{Training Pipeline.}  
First of all, we split the 2,000 annotated posts into a fixed development set (80\%) and a fixed test set (20\%). The development set is further split into training and validation sets using the same ratio but varying seeds. The distributions of intent labels for development set and test set are shown in Tab.~\ref{tab:incivility_distribution_dev_test}. It also statistically shows that the distribution of the six intents labels are similar in the development set and test set (see Appendix A).

\begin{table}[!htbp]
\centering
\caption{Distribution of intent labels across four categories of political incivility in the devlopment and test sets. Labels: 0=No defined labels, 1=Explicit, 2=Implicit, 3=Report, 4 = Intensify (no case); 5=Counter, 6=Escalate.}
\label{tab:incivility_distribution_dev_test}

\setlength{\tabcolsep}{3pt}
\renewcommand{\arraystretch}{1.15}

\begin{tabular}{lcccccc|cccccc}
\toprule
 & \multicolumn{6}{c|}{\textbf{Devlopment Set}} & \multicolumn{6}{c}{\textbf{Test Set}} \\
\cmidrule(lr){2-7} \cmidrule(lr){8-13}
\textbf{Form} & \textbf{0} & \textbf{1} & \textbf{2} & \textbf{3} & \textbf{5} & \textbf{6}
              & \textbf{0} & \textbf{1} & \textbf{2} & \textbf{3} & \textbf{5} & \textbf{6} \\
\midrule
IMP    & 89  & 268 & 24 & 5  & 9  & 5  & 22 & 67 & 6  & 2  & 2  & 1 \\
HSST   & 290 & 51  & 7  & 12 & 36 & 4  & 68 & 15 & 1  & 3  & 10 & 3 \\
PHAVPR & 292 & 8   & 8  & 53 & 31 & 8  & 70 & 2  & 2  & 17 & 5  & 4 \\
THREAT & 231 & 14  & 7  & 32 & 113 & 3  & 67 & 3  & 1  & 9  & 20 & 0 \\
\bottomrule
\end{tabular}

\end{table}

At each round of optimization, we first use instruction model to produce response for validation set and identity the targeted steps that make the most mistakes. The main hyperparameters of this process are summarized in Tab.~\ref{tab:hyperparams}.  

\begin{table}[!htbp]
\centering
\caption{Hyperparameters used in the training pipeline.}
\label{tab:hyperparams}

\setlength{\tabcolsep}{4pt} % tighter cols
\renewcommand{\arraystretch}{1.15}
\begin{tabularx}{\columnwidth}{l X}
\toprule
\textbf{Hyperparameter} & \textbf{Value} \\
\midrule
Instruction models & GPT-5-mini, Gemini 2.5 Flash-Lite, Claude 3 Haiku \\
Teacher/Refinement model & GPT-5-mini \\
Threshold for step optimization ($\tau$) & 0.1 \\
Ratio of validation set ($p$) & 0.2 \\
Optimization rounds ($T$) & 5 \\
\bottomrule
\end{tabularx}
\end{table}

We then run the optimization program only on the steps identified as problematic. The optimized strategies are selected based on validation performance and then evaluated on the held-out test set. After each round, we compute accuracy and weighted F1-score. We place greater emphasis on weighted F1 due to class imbalance. Improvements are reflected in higher accuracy and weighted F1-scores.

Prompting experiments were implemented through API calls to the official model endpoints using the \texttt{litellm} library. Fine-tuning was conducted on a single NVIDIA A100 SXM4 GPU (80~GB VRAM) with an AMD EPYC~7513 32-core CPU and 128~GB RAM, running CUDA~12.8.

\subsection{Main Results}
\paragraph{Overall comparison of LinGo with baselines.} 
Tab.~\ref{tab:comparison_zero_cot} reports the evaluation scores for zero-shot and CoT prompting across three models: GPT-5-mini, Gemini 2.5 Flash-Lite, and Claude 3 Haiku. Among them, GPT-5-mini achieves the highest performance on both accuracy and weighted F1, with scores of 0.583 and 0.578, respectively.

\begin{table}[!htbp]
\centering
\caption{Comparison across base prompting (Zero\textendash Shot) and Chain\textendash of\textendash Thought (CoT). Higher is better for Accuracy and weighted F1 (wF1).}
\label{tab:comparison_zero_cot}

\setlength{\tabcolsep}{6pt}
\renewcommand{\arraystretch}{1.15}

\begin{tabular}{@{}llcc@{}}
\toprule
\textbf{Model} & \textbf{Method} & \textbf{Accuracy} $\uparrow$ & \textbf{wF1} $\uparrow$ \\
\midrule
\multirow{2}{*}{GPT\textendash 5\textendash mini}
& Zero\textendash Shot & 0.518 & 0.513 \\
& CoT                  & \textbf{0.583} & \textbf{0.578} \\
\midrule
\multirow{2}{*}{Gemini 2.5\textendash Flash\textendash Lite}
& Zero\textendash Shot & 0.335 & 0.364 \\
& CoT                  & \textbf{0.380} & \textbf{0.413} \\
\midrule
\multirow{2}{*}{Claude 3 Haiku}
& Zero\textendash Shot & 0.190 & 0.130 \\
& CoT                  & \textbf{0.305} & \textbf{0.283} \\
\bottomrule
\end{tabular}

\end{table}

We then evaluate more advanced baselines using existing LLM optimization techniques, starting from the prompt without incorporating the linguistic-graph steps. Tab.~\ref{tab:tool_method_comparison} reports these results under the optimization setting for TextGrad, AdalFlow, DSPy, and RAG. Only with RAG (no linguistic graph added) do all models consistently improve: both accuracy and weighted F1 increase relative to the zero-shot and CoT baselines. In contrast, other optimization techniques sometimes perform worse than the zero-shot or CoT baselines. This may occur because some optimization methods introduce overfitting or incorporate misleading examples. Therefore, not all optimization techniques reliably improve classification performance. Among these tools, RAG achieves the strongest performance with Gemini 2.5 Flash-Lite (accuracy = 0.640; wF1 = 0.664). TextGrad yields the weakest performance and the smallest improvements. This suggests that optimizing demonstrations, for example, retrieving high-quality representative examples, is more effective than optimizing the text.

Finally, using the same models and the same LLM optimization techniques, we incorporate the linguistic graph to identify and refine only the targeted steps. As shown in the \textit{LinGO} section of Tab.~\ref{tab:tool_method_comparison}, Except for Claude 3 Haiku, LinGO improves almost all the accuracy and weighted F1 scores of LLMs compared to the direct optimization setting using any of the optimization techniques. Our approach is less effective on the Claude 3 Haiku model, likely due to its weaker semantic understanding and a higher risk of error propagation. However, it still works with TextGrad, and evaluation metrics are now all higher than the corresponding zero-shot and CoT baseline results. The best performance is again achieved by RAG, reaching an accuracy of 0.690 and a weighted F1 of 0.699. This corresponds to improvements of +10.7 pp (accuracy) and +12.1 pp (weighted F1) over the best zero-shot/CoT result, and +5.0 pp (accuracy) and +3.5 pp (weighted F1) over the best direct optimization result. It should be noted that techniques of metric-driven bootstrapping and similarity-based retrieval tend to be more consistent than textual gradient descent, as the latter depends on long-text generation and is therefore more variable.

\begin{table}[!htbp]
\centering
\caption{Performance comparison of prompt-optimization tools under two settings: (i) single-step \textit{Optimization} and (ii) step-wise LinGO. For each (Model, Method) block, the best tool by weighted F1 (wF1) is highlighted in bold. Acc.\ denotes accuracy.}
\label{tab:tool_method_comparison}

\setlength{\tabcolsep}{5pt}
\renewcommand{\arraystretch}{1.15}

\begin{tabular}{@{}lllcc@{}}
\toprule
\textbf{Model} & \textbf{Method} & \textbf{Tool} & \textbf{Acc.} $\uparrow$ & \textbf{wF1} $\uparrow$ \\
\midrule

\multirow{9}{*}{GPT-5-mini}
& \multirow{4}{*}{Direct Optimization}
& TextGrad             & 0.328 & 0.393 \\
&                      & AdalFlow            & 0.568 & 0.411 \\
&                      & DSPy                & 0.540 & 0.525 \\
&                      & RAG                 & 0.638 & 0.640 \\
\cmidrule(lr){2-5}
& \multirow{5}{*}{LinGO}
& TextGrad             & 0.590 & 0.614 \\
&                      & AdalFlow            & 0.620 & 0.637 \\
% &                      & DSPy (full\_optimization) & -- & -- \\
&                      & DSPy                & 0.508 & 0.554 \\
&                      & RAG                 & \textbf{0.655} & \textbf{0.677} \\
\midrule

\multirow{9}{*}{Gemini 2.5 Flash\textendash Lite}
& \multirow{4}{*}{Direct Optimization}
& TextGrad             & 0.413 & 0.392 \\
&                      & AdalFlow            & 0.493 & 0.427 \\
&                      & DSPy                & 0.630 & 0.606 \\
&                      & RAG                 & 0.640 & 0.664 \\
\cmidrule(lr){2-5}
& \multirow{5}{*}{LinGO}
& TextGrad             & 0.518 & 0.550 \\
&                      & AdalFlow            & 0.530 & 0.551 \\
% &                      & DSPy (full\_optimization) & -- & -- \\
&                      & DSPy                & 0.648 & 0.665 \\
&                      & RAG                 & \textbf{0.690} & \textbf{0.699} \\
\midrule

\multirow{9}{*}{Claude 3 Haiku}
& \multirow{4}{*}{Direct Optimization}
& TextGrad             & 0.203 & 0.257 \\
&                      & AdalFlow            & 0.488 & 0.385 \\
&                      & DSPy                & \textbf{0.515} & 0.540 \\
&                      & RAG                 & 0.495 & \textbf{0.551} \\
\cmidrule(lr){2-5}
& \multirow{5}{*}{LinGO}
& TextGrad             & 0.350 & 0.348 \\
&                      & AdalFlow            & 0.365 & 0.380 \\
% &                      & DSPy (full\_optimization) & -- & -- \\
&                      & DSPy                & 0.410 & 0.458 \\
&                      & RAG                 & 0.479 & 0.489 \\
\bottomrule
\end{tabular}

\end{table}

The experimental results show that our approach, namely prompting LLMs based on linguistic structures and iteratively optimizing the targeted steps, improves their ability to capture complex semantics and achieves the best performance in terms of accuracy and weighted F1 compared to zero-shot/CoT prompting and direct optimization.

Besides, we also evaluate our method against fine-tuned open-source models (see Tab.~\ref{tab:opensource_vs_lingo}). Experiments with three models - Mistral-7B, DeepSeek-V2-Lite-Chat, and Qwen3-4B-Instruct-2507 show that the best results reach 0.590 accuracy and 0.535 weighted F1, both lower than our method’s best performance with GPT-5-mini and Gemini 2.5 Flash-Lite. Performance may improve with larger open-source models, but this would require substantially greater computational resources and infrastructure costs.

\begin{table}[!htbp]
\centering
\caption{Comparison between fine-tuned open-source models with human annotations and closed-source models optimized with LinGO. LinGO results are the best round by weighted F1 score, with model Gemini 2.5 Flash–Lite and RAG.}
\label{tab:opensource_vs_lingo}

\begin{tabular}{@{}lccc@{}}
\toprule
\textbf{Model} & \textbf{Method} & \textbf{Accuracy} $\uparrow$ & \textbf{wF1} $\uparrow$ \\
\midrule
Mistral--7B & FT & 0.215 & 0.250 \\
DeepSeek--V2--Lite--Chat & FT & 0.302 & 0.340 \\
Qwen3--4B--Instruct--2507 & FT & \textbf{0.590} & \textbf{0.535} \\
\midrule
Gemini 2.5 Flash--Lite & LinGO (RAG) & \textbf{0.690} & \textbf{0.699} \\
\bottomrule
\end{tabular}

\end{table}

\begin{figure}[!htbp]
  \centering
  \includegraphics[width=\linewidth]{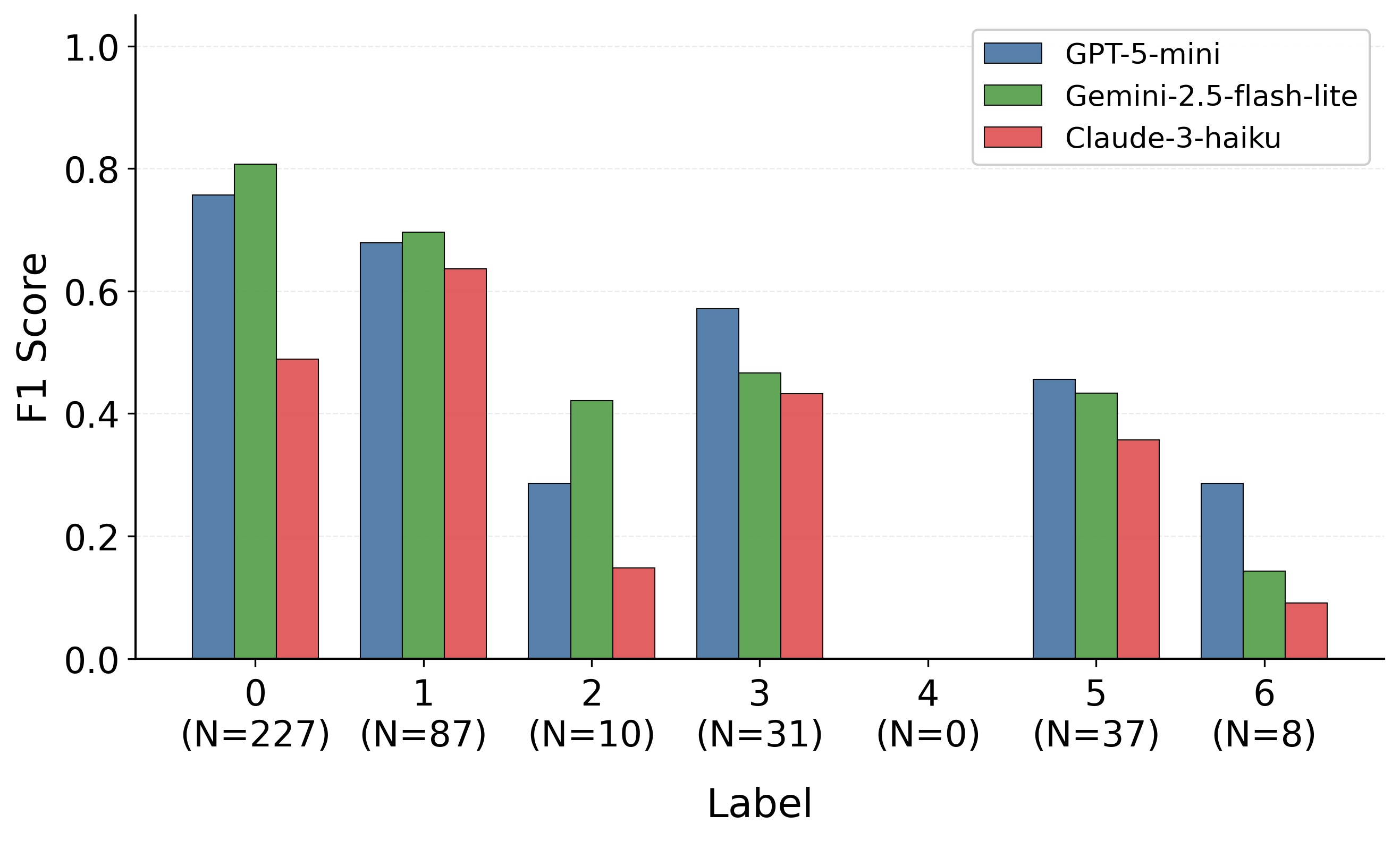} 
  \caption{Comparison of LinGO and baseline prompting methods across intent labels (0-6) and models (GPT-5-mini, Claude~3~Haiku, Gemini~2.5~Flash–Lite).}
  \label{fig:metrics_per_label}
\end{figure}

\paragraph{Performance on different intent labels across models.} We further analyze performance (based on F1 score) across intent labels for the three models under our approach with RAG, and observe several common patterns. As shown in Fig.~\ref{fig:metrics_per_label}, the label distribution in our test set is notably imbalanced; for example, there are no instances of the “intensifying” category (label~4). Among the remaining labels, direct incivility achieves higher F1 scores than indirect incivility, which is not surprising due to the difficulty of interpretating indirect incivility. Among indirect expressions, implicit incivility is particularly difficult to detect, which has also been mentioned by prior work \cite{stoll2020detecting, park2024collaborative, rottger2020hatecheck}.

Across models, Claude 3 Haiku performs worse on all labels than GPT-5-mini and Gemini 2.5 Flash–Lite. Gemini 2.5 Flash–Lite is slightly better than GPT-5-mini at identifying irrelevant cases (label~0), direct incivility (label~1), and implicit incivility (label~2), whereas GPT-5-mini performs better on referred indirect incivility, including label~3 (Reporting), label~5 (Criticizing) and label~6 (Escalating). These differences suggest that practitioners may benefit from benchmarking multiple models and selecting those that perform best on the labels most relevant to their research goals.

% GPT–5-mini demonstrates strong baseline performance in distinguishing civil from uncivil discourse when explicit indicators are present. The main performance gains from our approach appear in label~5 (Countering) and label~6 (Escalating), which involve more linguistically complex intents. In contrast, Claude~3~Haiku tends to overestimate uncivil content, resulting in lower baseline accuracy for civil cases but better recognition of indirect incivility compared to GPT–4o. After applying our approach, the model becomes more conservative—many uncivil cases are reassigned to label~0—leading to little or even reduced improvement in identifying indirect incivility. Gemini~2.5~Flash–Lite performs intermediately between GPT–4o and Claude~3~Haiku in distinguishing civil and indirect expressions of incivility. While our approach enhances its ability to separate civil and directly uncivil cases, it still struggles with indirect ones.

% Although models—Claude~3~Haiku and Gemini~2.5~Flash–Lite achieve a higher overall improvement than GPT–4o, they show limited progress in recognizing indirect categories. This finding suggests that when the research goal is to differentiate subtle and indirect intents, our method performs best with advanced models that are already capable of detecting explicit uncivil cues. Moreover, it underscores the importance of evaluating which components of a multi-step optimization pipeline are being improved, rather than focusing solely on aggregate performance metrics.

\begin{figure}[!htbp]
  \centering
  \includegraphics[width=\linewidth]{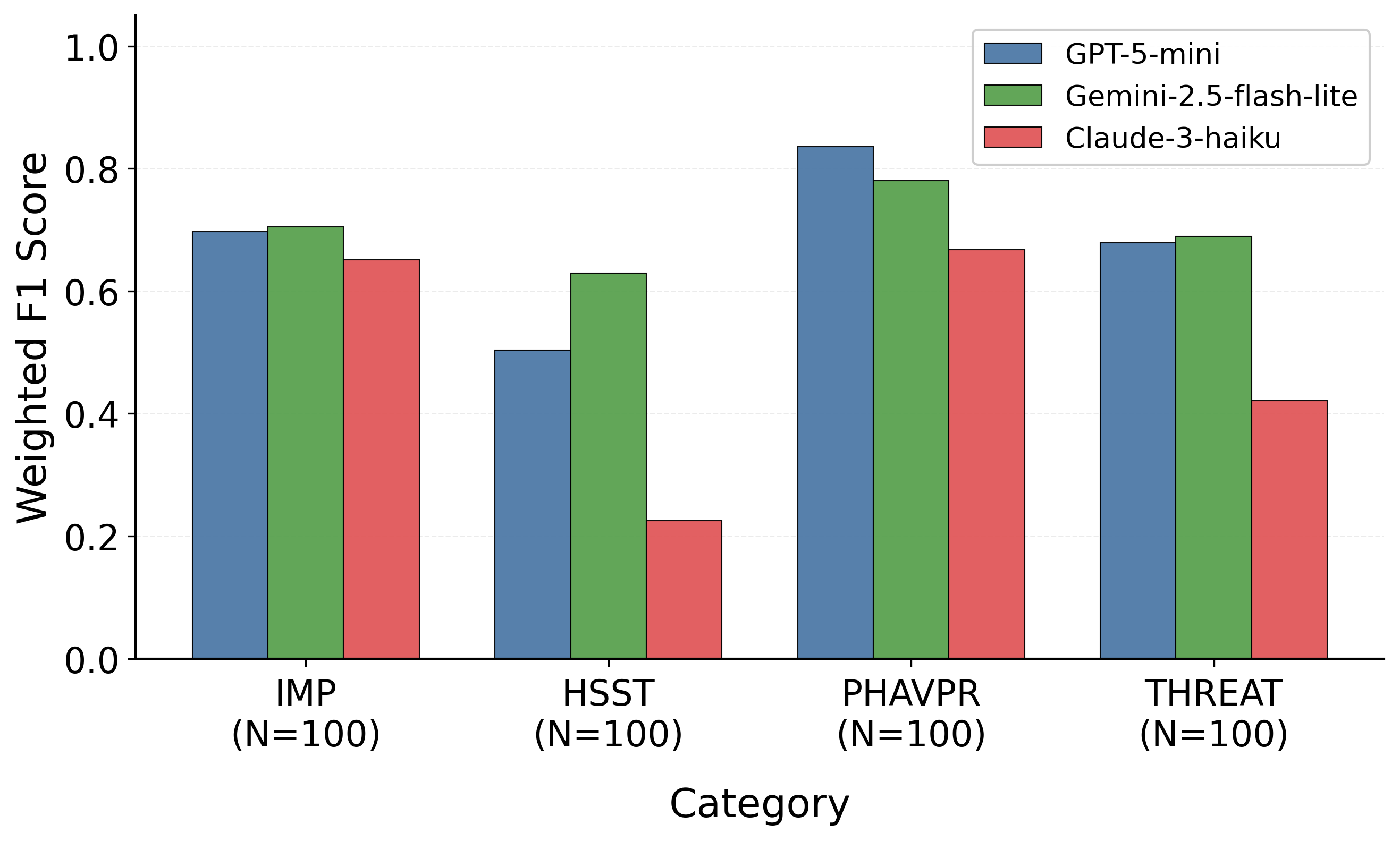} 
  \caption{Comparison of LinGO and baseline prompting methods across forms of incivility (IMP, HSST, PHAVPR, THREAT) and models (GPT-5-mini, Claude~3~Haiku, Gemini~2.5~Flash–Lite).}
  \label{fig:metrics_per_form}
\end{figure}

\paragraph{Performance on different categories of incivility across models.}
We further examine the weighted F1 scores of the three models under the RAG setting across the four categories of incivility (Fig.~\ref{fig:metrics_per_form}). A consistent pattern emerges across models: \emph{Hate Speech and Stereotyping} yields the lowest weighted F1, whereas \emph{Physical Harm and Violent Political Rhetoric} achieves the highest.

Across all categories, Claude 3 Haiku performs worst relative to GPT-5-mini and Gemini 2.5 Flash–Lite. GPT-5-mini and Gemini 2.5 Flash–Lite achieve broadly similar performance overall, but Gemini 2.5 Flash–Lite performs slightly better on \emph{Hate Speech and Stereotyping} than GPT-5-mini, and GPT-5-mini is better in recognizing \emph{Physical Harm and Violent Political Rhetoric}. This may reflect differences in their training data coverage or model sensitivity to hate speech–related patterns. Therefore, when selecting a model, practitioners should consider which incivility categories they aim to detect and choose the model that previously performs better on similar category tasks.

% the largest improvement for GPT–4o occurs in the category \emph{Threats to Democratic Institutions and Values}. This aligns with our observation that GPT–4o primarily improves in detecting indirect incivility, which is most frequently found in this category. In contrast, Claude 3 Haiku and Gemini 2.5 Flash–Lite exhibit greater improvements in \emph{Hate Speech and Stereotyping}, \emph{Physical Harm and Violent Political Rhetoric}, and \emph{Threats to Democratic Institutions and Values}, as these categories include a higher proportion of label~0 cases (see Tab.~\ref{tab:incivility_distribution_dev_test}). For Claude~3~Haiku, although our approach substantially enhances performance across three forms of incivility, it reduces accuracy in \emph{Impoliteness}.

% These findings suggest that different forms of incivility may reveal distinct types of model bias. Future studies should adopt a multidimensional definition of incivility and apply models tailored to each dimension to improve automatic detection.

\section{Conclusion}
Previous approaches to the automatic detection of incivility often struggle to distinguish between the direct expression of incivility and indirect mentions of it. Although both contain uncivil features, they may convey different intents depending on how they are expressed. Our work identifies six distinct expressions and intents by introducing a linguistic graph  optimization framework. The results show that our method outperforms zero-shot learning, CoT prompting, the state-of-the-art LLM optimization techniques, as well as some fine-tuned open-source models. A deeper analysis of performance by label and incivility category suggests that different models excel at different aspects of the task. For instance, GPT-5-mini performs better at detecting referred indirect labels, whereas Gemini-2.5-Flash-Lite performs superbly at recognizing hate speech and stereotyping.

Our approach provides a practical method for LLMs to explain fine-grained direct and indirect forms of incivility, and makes the sub-steps of explanation explicit and optimizable. It can be used by other researchers either as a straightforward classifier for multi-label incivility intent classification or as a post hoc analysis tool for uncivil cases identified by other machine learning or deep learning classifiers. Identifying various linguistic structures and its underlying intents can inform more accurate examination of harmfulness and interventions to mitigate toxic online environments.

\section{Limitations}
Our approach has three main limitations. First, it depends on human annotation not only for intent labels but also for intermediate sub-step labels at each linguistic node, which increases labeling effort. However, our experiments show that using only 2,000 labeled instances (500 for single category) has been enough to improve LLM classification performance. Second, because the method iteratively refines prompts and evaluates them on full datasets, it can be computationally demanding. Our experiments use low-cost models, keeping the cost at \$5–\$10 per experiment, and the highest weighted F1 score is already close to 70\%, which is quite acceptable. Third, we evaluate the approach on a single domain (Brazilian politics), language (Portuguese), and four categories of uncivil discourse. Generalizability cannot be fully confirmed, and we encourage future work to test the method across additional cultural contexts, languages, and incivility types.

\section{Ethical considerations}
Automatic detection of uncivil intents may be misused for surveillance or censorship and could compromise individual privacy. We recommend combining automated detection results with human judgment when implementing moderation measures. 

This study uses only publicly available Twitter/X posts. To protect user privacy, all user identifiers were removed to prevent re-identification. Because the research focuses on online incivility, the dataset includes some offensive language; such content was analyzed exclusively for scholarly purposes. No personally identifiable or harmful text is reproduced in this paper, and all data handling procedures complied with ethical research standards and were approved by the University of Zurich Ethics Committee.

We affirm that AI-assisted tools were used solely to correct grammatical errors and enhance the clarity and readability of the manuscript. All ideas, analyses, and interpretations are entirely those of the authors.

\section{Code and Data Availability}
The code and data sets for this work is available at \url{https://github.com/yuanzhang1227/Replication_Code_LinGO}. 

\appendix
\section*{Appendix}
\section{Label distribution in the development and test sets}

\begin{figure}[H]
  \centering
  \includegraphics[width=0.8\linewidth]{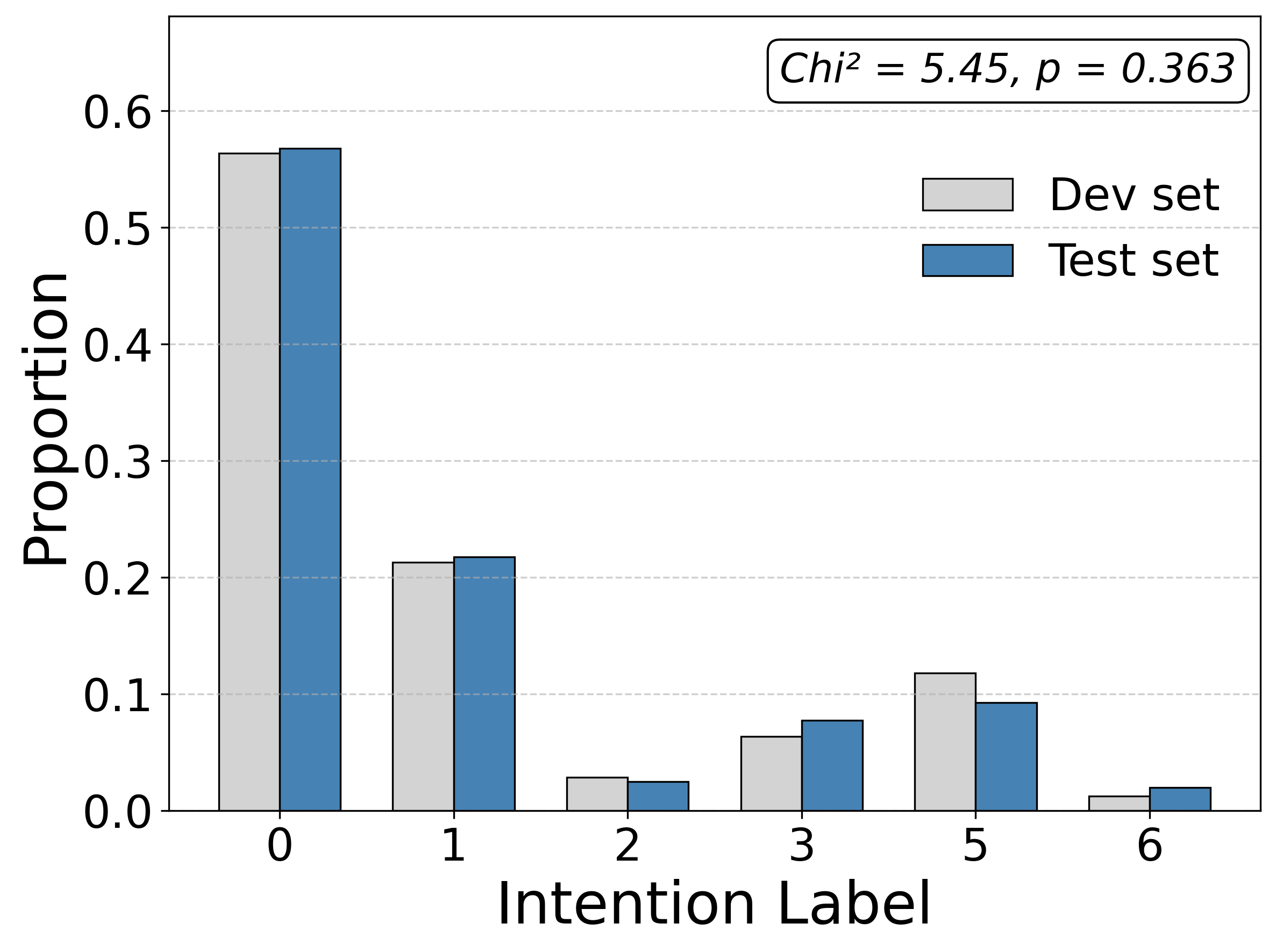} 
  \caption{Distribution of intent labels in the development and test sets. The chi-square test shows that the differences between their distributions are not statistically significant.}
  \label{fig:distribution}
\end{figure}

\section{Formalization of LinGO Process}
\subsection{Linguistic Graph Construction}
\begin{figure}[h]
  \centering
  \includegraphics[width=1.0\linewidth]{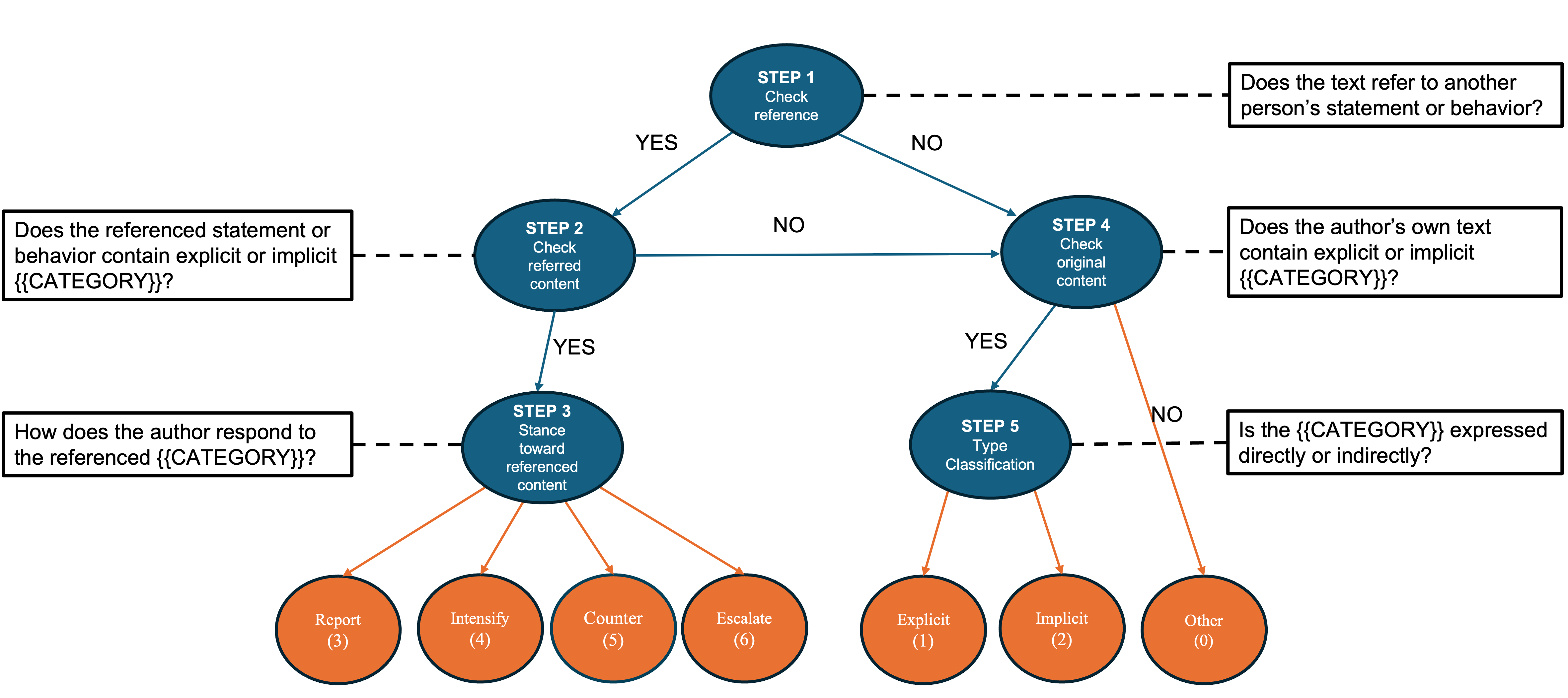}
  \caption{Linguistic Graph Based on Discourse Representation.}
  \label{fig:linguistic graph}
\end{figure}

An illustrative graph of LinGO is given in Fig.~\ref{fig:linguistic graph}.

Formally, we define these different steps as a directed acyclic graph 
\[
\mathcal{T} = (\mathcal{N}, \mathcal{E}),
\]
where $\mathcal{N}$ is the set of linguistic nodes (reasoning steps for LLMs) and $\mathcal{E} \subseteq \mathcal{N} \times \mathcal{N}$ is the set of directed edges that encode the flow of reasoning. Each node $n \in \mathcal{N}$ is associated with a decision function
\[
d_n : \mathcal{X} \to \mathcal{A}_n,
\]
where $\mathcal{X}$ is the input space and $\mathcal{A}_n$ is the set of possible answers at step $n$ (e.g., \{YES, NO\}, or stance categories).  

The transition rule is then defined as
\[
\delta : \mathcal{N} \times \mathcal{A}_n \to \mathcal{N},
\]
which maps a node $n$ and an answer $a \in \mathcal{A}_n$ to the next node in the graph.  

Reasoning on input $x \in \mathcal{X}$ proceeds as a path
\[
\pi(x) = (n_1, a_1), (n_2, a_2), \dots, (n_m, a_m),
\]
where $n_1$ is the initial node of the graph, $a_i = d_{n_i}(x)$ (d is the process of LLM generation), and $n_{i+1} = \delta(n_i, a_i)$.  

Finally, the leaf node $n_m$ corresponds to a classification function
\[
f_{n_m} : \mathcal{X} \to \mathcal{Y},
\]
which outputs a label $y \in \mathcal{Y}$ from the pre-defined label set. In our study, they are different intents of uncivil expression: explicit, implicit, report, intensify, counter, and escalate.    

\subsection{Prompt Optimization via Step-wise Refinement}

We first construct an initial prompt $\mathcal{P}^{(0)}$ following the nodes in the linguistic graph
$\mathcal{T}=(\mathcal{N},\mathcal{E})$, where each node $n\in\mathcal{N}$ corresponds to a sub-decision and edges
encode the branching logic. The prompt contains (i) category definitions; (ii) main task description, (iii) sub-tasks description, and (iv) output format. Given an input $x\in\mathcal{X}$, executing $\mathcal{P}^{(t)}$ induces
(i) a \emph{reasoning path} $\pi^{(t)}(x)$, i.e., the sequence of visited nodes and decisions, and (ii) a terminal
\emph{label} $\hat y^{(t)}(x)\in\mathcal{Y}$. See examples of the prompt:

% Box 1: Category Definitions
\begin{tcolorbox}[title=\textbf{(i) Category Definitions}, colback=blue!5!white, colframe=blue!75!black, boxsep=2pt, left=4pt, right=4pt, top=2pt, bottom=2pt, fontupper=\small]
Always taken from the first square-bracket tag (e.g., [Impoliteness], [Hate Speech and Stereotyping], [Physical Harm and Violent Political Rhetoric], [Threats to Democratic Institutions and Values]).

\textbf{Impoliteness:} messages including rudeness/disrespect (name-calling, aspersions, calling someone a liar, hyperbole, non-cooperation, pejoratives, vulgarity, belittling, shouting via ALL-CAPS or excessive ``!'' when context supports it).
\textbf{Hate Speech and Stereotyping:} discriminatory content targeting protected groups (gender identity, sexual orientation, religion, race, nationality, ideology, disability); over-generalizations, out-group demeaning; (do NOT include purely individual appearance insults unless tied to group identity).
\textbf{Physical Harm and Violent Political Rhetoric:} threats/advocacy/praise of physical harm or violence; direct or metaphorical calls for harm; justification of violence for political ends.
\textbf{Threats to Democratic Institutions and Values:} advocacy or approval of actions undermining elections/institutions/rule of law/press freedom/civil rights; promotion of autocracy; unfounded claims that delegitimize institutions.
\end{tcolorbox}

\vspace{-0.5em}

% Box 2: Main Task Description
\begin{tcolorbox}[title=\textbf{(ii) Main Task Description}, colback=green!5!white, colframe=green!75!black, boxsep=2pt, left=4pt, right=4pt, top=2pt, bottom=2pt, fontupper=\small]
You are an annotator interpreting the intent of uncivil Portuguese social media discourse about Brazilian politics on X (formerly Twitter). Analyze the intent behind the given post and assign the most appropriate intent label from the list below (apply ONLY to the target category from the tag):
1 = Explicit \{\{CATEGORY\}\}: direct, overt \{\{CATEGORY\}\}.
2 = Implicit \{\{CATEGORY\}\}: indirect, veiled \{\{CATEGORY\}\}.
3 = Report \{\{CATEGORY\}\}: quotes/refers to \{\{CATEGORY\}\} content without opinion.
4 = Intensify \{\{CATEGORY\}\}: quotes/refers to \{\{CATEGORY\}\} content and agrees/amplifies.
5 = Counter \{\{CATEGORY\}\}: quotes/refers to \{\{CATEGORY\}\} content and criticizes/disagrees.
6 = Escalate \{\{CATEGORY\}\}: responds to \{\{CATEGORY\}\} content with \{\{CATEGORY\}\}.
0 = Does not fit any of the above patterns.
\end{tcolorbox}

\vspace{-0.5em}

% Box 3: Sub-tasks Description (Decision Process)
\begin{tcolorbox}[title=\textbf{(iii) Sub-tasks Description (Linguistic Graph)}, colback=orange!5!white, colframe=orange!75!black, boxsep=2pt, left=4pt, right=4pt, top=2pt, bottom=2pt, fontupper=\small]
\textbf{STEP 1: Check Reference.} Question: Does the text refer to another person's statement or behavior? If NO $\rightarrow$ go to STEP 4. If YES $\rightarrow$ go to STEP 2.

\textbf{STEP 2: Check Referenced Content.} Question: Does the referenced statement or behavior contain explicit or implicit \{\{CATEGORY\}\}? If NO $\rightarrow$ go to STEP 4. If YES $\rightarrow$ go to STEP 3.

\textbf{STEP 3: Stance Toward Referenced Content.} Question: How does the author respond to the referenced \{\{CATEGORY\}\}? Report (3): mentions without opinion. Intensify (4): agrees or amplifies. Counter (5): criticizes or disagrees. Escalate (6): responds to \{\{CATEGORY\}\} content with \{\{CATEGORY\}\}.

\textbf{STEP 4: Check Original Content.} Question: Does the author's own text contain explicit or implicit \{\{CATEGORY\}\}? If NO $\rightarrow$ Label 0. If YES $\rightarrow$ go to STEP 5.

\textbf{STEP 5: Type Classification.} Question: Is the \{\{CATEGORY\}\} expressed directly or indirectly? Explicit (1): direct, overt \{\{CATEGORY\}\}. Implicit (2): indirect, veiled \{\{CATEGORY\}\}.
\end{tcolorbox}

\vspace{-0.5em}

% Box 4: Output Format
\begin{tcolorbox}[title=\textbf{(iv) Output Format}, colback=red!5!white, colframe=red!75!black, boxsep=2pt, left=4pt, right=4pt, top=2pt, bottom=2pt, fontupper=\small]
Return ONLY valid JSON: \texttt{\{"LABEL": <int>, "REASONING": \{"STEP 1": "YES"/"NO", ...\}\}}

\textbf{Examples:}
(1) STEP 1=NO $\rightarrow$ STEP 4=YES $\rightarrow$ STEP 5: \texttt{\{"LABEL": 1, "REASONING": \{"STEP 1": "NO", "STEP 4": "YES", "STEP 5": "Explicit"\}\}}
(2) STEP 1=YES $\rightarrow$ STEP 2=YES $\rightarrow$ STEP 3: \texttt{\{"LABEL": 5, "REASONING": \{"STEP 1": "YES", "STEP 2": "YES", "STEP 3": "Counter"\}\}}
(3) STEP 1=YES $\rightarrow$ STEP 2=NO $\rightarrow$ STEP 4=NO: \texttt{\{"LABEL": 0, "REASONING": \{"STEP 1": "YES", "STEP 2": "NO", "STEP 4": "NO"\}\}}
(4) STEP 1=NO $\rightarrow$ STEP 4=NO: \texttt{\{"LABEL": 0, "REASONING": \{"STEP 1": "NO", "STEP 4": "NO"\}\}}
\end{tcolorbox}

\paragraph{Data partitioning.}
Given a labeled dataset $\mathcal{D}\subseteq\mathcal{X}\times\mathcal{Y}$ with gold terminal labels $y^\ast(x)$ and gold node answers $\{a_n^\ast(x)\}_{n\in\mathcal{N}}$, we partition the development data as
\[
\mathcal{D}_{\mathrm{dev}} \;=\; \mathcal{D}_{\mathrm{train}} \;\cup\; \mathcal{D}_{\mathrm{val}},
\]
and reserve a held-out test set $\mathcal{D}_{\mathrm{test}}$ for a single final evaluation.

\paragraph{Diagnosis on validation set.}
At optimization round $t$, we run the current configuration (prompt/program, and retrieved demonstrations if used)
on each $x\in\mathcal{D}_{\mathrm{val}}$ to obtain the predicted label $\hat y^{(t)}(x)$ and node-level answers
$\{\hat a_n^{(t)}(x)\}_{n\in\mathcal{N}}$ along the realized path $\pi^{(t)}(x)$. We define the set of validation
instances that visit node $n$ as
\[
\mathcal{I}_n^{(t)} \;=\; \{\,x\in\mathcal{D}_{\mathrm{val}}:\; n\in\pi^{(t)}(x)\,\}.
\]

\paragraph{Step-wise mismatch estimation on validation data.}
For each visited node $n$, we compute the mismatch indicator
\[
M_n^{(t)}(x) \;=\; \mathbf{1}\!\left\{\,\hat a_n^{(t)}(x)\neq a_n^\ast(x)\,\right\},
\]
and the empirical mismatch rate
\[
\hat p_n^{(t)} \;=\; \frac{1}{|\mathcal{I}_n^{(t)}|}\sum_{x\in\mathcal{I}_n^{(t)}} M_n^{(t)}(x),
\]
restricting the sum to instances where $\hat y^{(t)}(x)\neq y^\ast(x)$. 

\paragraph{Target selection.}
We select nodes whose mismatch rates exceed a threshold $\tau$:
\[
\mathcal{S}^{(t)} \;=\; \{\, n\in\mathcal{N}:\; \hat p_n^{(t)}>\tau \,\}.
\]
If $\mathcal{S}^{(t)}=\varnothing$, the procedure terminates early.

\paragraph{Targeted updates using training data.}
For each selected node $n\in\mathcal{S}^{(t)}$, we construct a node-specific training signal from
$\mathcal{D}_{\mathrm{train}}$, optionally prioritized according to error patterns observed on
$\mathcal{D}_{\mathrm{val}}$, and apply an optimizer
\[
\omega_n \in \Omega=\{\textsc{DSPy},\textsc{TextGrad},\textsc{AdalFlow},\textsc{RAG}\}
\]
to update only the components associated with node $n$ (e.g., step instruction text, step-specific programs/demos, or
retrieval policy). All non-selected nodes remain fixed.

\paragraph{Validation-based model selection.}
After each round, we evaluate the updated configuration on $\mathcal{D}_{\mathrm{val}}$ using a metric
$m_{\mathrm{val}}$ (e.g., weighted F1) and retain the best checkpoint:
\[
(\mathcal{P}^\star,\mathcal{F}^\star) \;=\; \arg\max_t \; m_{\mathrm{val}}\!\big(\mathcal{P}^{(t)},\mathcal{F}^{(t)},\mathcal{D}_{\mathrm{val}}\big).
\]

\paragraph{Final evaluation.}
We report final performance by evaluating $(\mathcal{P}^\star,\mathcal{F}^\star)$ on
$\mathcal{D}_{\mathrm{test}}$.

\bibliographystyle{splncs04}
\bibliography{custom}

\end{document}